\documentclass{article}

\PassOptionsToPackage{numbers, compress}{natbib}

\usepackage[preprint]{neurips_2024}

\usepackage[utf8]{inputenc} 
\usepackage[T1]{fontenc}    
\usepackage{hyperref}       
\usepackage{url}            
\usepackage{booktabs}       
\usepackage{amsfonts}       
\usepackage{nicefrac}       
\usepackage{microtype}      
\usepackage{xcolor}         
\usepackage{graphicx}
\usepackage{amsmath}
\usepackage{multirow}
\usepackage{subcaption}
\usepackage{makecell}
\usepackage{tabularx}
\usepackage{wrapfig}

\title{OutlierTune: Efficient Channel-Wise Quantization for Large Language Models}

\author{%
  Jinguang Wang$^{1,2}$, Yuexi Yin$^{1}$, Haifeng Sun$^{1*}$, Qi Qi$^{1}$, Jingyu Wang$^{1}$, Zirui Zhuang$^{1}$,\\ \textbf{Tingting Yang$^{2}$, Jianxin Liao$^{1}$,}  \\
  $^1$Beijing University of Posts and Telecommunications,\\
 $^2$PengCheng Laboratory\\
}

\begin{document}

\maketitle

\begin{abstract}
Quantizing the activations  of large language models (LLMs) has been a significant challenge due to the presence of structured outliers. Most existing methods focus on the per-token or per-tensor quantization of activations, making it difficult to achieve both accuracy and hardware efficiency. To address this problem, we propose OutlierTune, an efficient per-channel post-training quantization (PTQ) method for the activations of LLMs. OutlierTune consists of two  components: pre-execution of dequantization and symmetrization. The pre-execution of dequantization updates the model weights by the activation scaling factors, avoiding the internal scaling and costly additional computational overheads brought by the per-channel activation quantization. The symmetrization further reduces the quantization differences arising from the weight updates by ensuring the balanced numerical ranges across different activation channels. OutlierTune is easy to implement and hardware-efficient, introducing almost no additional computational overheads during the inference. Extensive experiments show that the proposed framework outperforms existing methods across multiple different tasks. Demonstrating better generalization, this framework improves the Int6 quantization of the instruction-tuning LLMs, such as OPT-IML, to the same level as half-precision (FP16). Moreover, we have  shown that the proposed framework is 1.48$\times$ faster than the FP16 implementation while reducing approximately 2$\times$ memory usage. 

\end{abstract}

\section{Introduction}
Large language models (LLMs) have attracted considerable attention and demonstrated remarkable performance in various tasks \cite{vaswani2017attention,touvron2023llama,radford2019language,NEURIPS2022_77c6ccac, workshop2022bloom, zhang2022opt, brown2020language}, but their  actual deployments have been hindered due to the significant memory and computational overheads \citep{liu2023scissorhands,pope2022efficiently}. As an effective method to alleviate the deployment difficulties of LLMs,  quantization can effectively reduce the computational and storage costs  by converting the model  parameters  into the fixed-point or low-precision floating-point representations \cite{wang2019haq,tao-etal-2022-compression,zhang-etal-2020-ternarybert,bai-etal-2021-binarybert,pmlr-v139-kim21d}.

Quantization presents an appealing solution for the deployment of LLMs, yet its practical implementation poses challenges. Quantization-aware training (QAT) \cite{bengio2013estimating,gholami2021survey,choi2018pact}, which quantizes models during training, offers significant improvements, but isn't always practical for LLMs due to the vast number of parameters involved. Compared to QAT, post-training quantization (PTQ) \cite{polino2018model,Jacob_2018_CVPR} is an easier-to-implement scheme, simplifying the process by eliminating the expensive training prerequisites \cite{frantar2022gptq,lin2023awq}, but performing low-bit PTQ on the weights and activations of LLMs presents a formidable challenge, often resulting in a significant degradation in model performance \cite{wei-etal-2023-outlier}. This degradation is primarily due to the structured outliers in certain fixed activation channels \cite{dettmers2022llmint8}, which serve as the primary source of quantization errors during the low-bit activation quantizations.

To address the challenge posed by the outlier channels during quantization, some works have dedicated to reduce  their impacts on the quantization scale. LLM.int8 \cite{dettmers2022llmint8} effectively isolates the outliers in activation quantization by preserving the outlier channels at FP16 precision.  SmoothQuant \citep{pmlr-v202-xiao23c} employs a smoothing technique to prepare activations for easier quantization. Outlier Suppression \citep{wei2022outlier} narrows the numerical range of outlier channels to reduce the quantization errors. The above methods focus on quantizing activations in the token or tensor dimension, which inevitably introduces significant quantization errors when operating at low precision.

Recent studies have shown that per-channel activation quantization can effectively mitigate quantization errors by reducing the influence of outlier channels on the quantization scale \citep{pmlr-v202-xiao23c,wu2023understanding}.
 However, the per-channel activation quantization faces significant hurdles during the dequantization process. A primary reason  is  the scaling inside matrix multiplications caused by the per-channel quantization, which not only introduces considerable computational overheads but also poses difficulties in aligning with hardware-accelerated kernels. To address this challenge, several methods have been developed to reduce the additional  computational overhead of per-channel quantization by utilizing simple reordering techniques \cite{yuan2023rptq,wu2023understanding}.
 However, such methods still cannot simultaneously satisfy the efficient hardware implementation and have poor latency performance. In comparison, SPIQ \cite{yvinec2023spiq} implements the per-channel activation quantization in a static quantization manner, but its applicability is limited to smaller vision models such as MobileNet and DenseNet.

In this paper, we introduce OutlierTune, a novel weight-activation quantization framework, which enables efficient inference while maintaining the accuracy of per-channel activation quantization. OutlierTune incorporates two pivotal components: pre-execution of dequantization and symmetrization.
 The pre-execution of dequantization updates the weights by the activation scaling factors and avoids the internal scaling of dequantization by a mathematically equivalent transformation. This optimizes the per-channel activation quantization process and improves the model latency performance. Then to further improve accuracy, we employ the symmetrization to balance the numerical ranges across different activation channels.
 This adjustment reduces the adverse effects of the activation scaling factor on weight quantization and fosters the quantization-friendly activations.
OutlierTune is easy to implement and hardware-efficient. Experimental results demonstrate its ability to quantize  the model weights and activations to 6 bits without significant accuracy loss.

The proposed OutlierTune framework shows the efficient implementation of per-channel activation quantization on LLMs: prior works resulted in poor latency performance or were not suitable for LLMs \cite{yuan2023rptq, wu2023understanding, yvinec2023spiq}. The pre-execution of dequantization and symmetrization in OutlierTune can be accomplished beforehand by tuning the weights and biases, thus avoiding the additional computations during the inference. Moreover, OutlierTune eliminates the need for tedious parameter search process.

The main contributions  are outlined as follows:
\begin{itemize}
\item  {We study in detail the difficulty of quantizing activations along the channel dimension and propose our OutlierTune framework including the pre-execution of dequantization and symmetrization. This framework is simple and efficient, and can be plug-and-play.}
\item  {The pre-execution of dequantization optimizes the per-channel activation
quantization  process  by eliminating internal scaling and reducing computational overheads. Meanwhile, the symmetrization harmonizes  the numerical ranges of outlier channels to mitigate the quantization errors of updated weights.}
\item  {We validate the effectiveness of our OutlierTune framework across different LLMs.  The evaluation  results across different tasks demonstrate that OutlierTune does not incur significant model performance loss when operating under 8-bits and 6-bits quantizations. Compared to FP16, OutlierTune can achieve up to 1.48$\times$ inference speed-up and approximately 2$\times$ memory saving.}
\end{itemize}

\section{Preliminaries}

\subsection{Quantization}
Quantization reduces memory usage and increases computational efficiency by converting high-precision floating-point to fixed-point representation \cite{Jacob_2018_CVPR}. Governed by scaling factor $s$ and zero point $z$, quantization can be mathematically expressed as:
\begin{equation}\label{a}
\begin{aligned}
\tilde x =\text{clip}\left(\left\lceil \frac{x}{s} \right\rfloor + z, 0, 2^b\!-\!1 \right), \ \ \ 
\bar x = ( \tilde x - z) s,
\end{aligned}
\end{equation}
where $x$ is the floating-point tensor,  $\tilde x$ is the fixed-point representation, $\bar x$ is the inverse quantized counterpart, $b$ is the quantization bits, $\lceil \rfloor$ denotes the rounding operation. Quantization techniques can be categorized into symmetric and asymmetric types based on the position of the zero point $z$.  Symmetric quantization, centered on the zero point $(z=0)$, is well-suited for processing data with evenly distributed positive and negative numbers \citep{pmlr-v202-xiao23c}. In contrast, asymmetric quantization permits a flexible zero point $(z\neq 0)$, which allows for better adaptation to various data distributions \cite{frantar2022gptq}.


\begin{figure}[tbp]
\center{\includegraphics[width=14.0cm]  {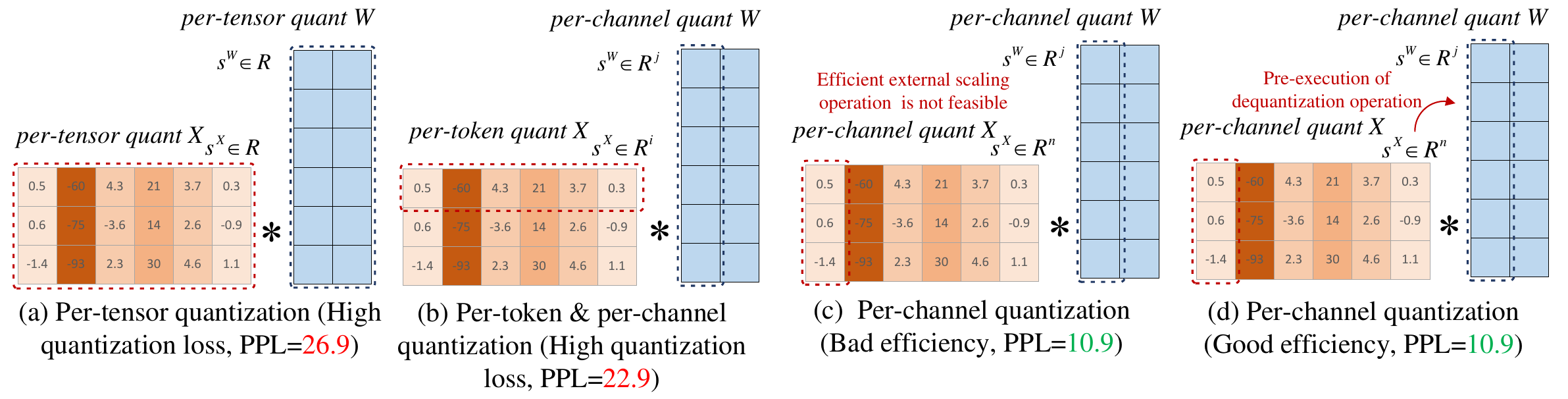}}
\caption{\label{fig1} Differences in different quantization dimensions. The quantization settings in Fig. \ref{fig1}a,\ref{fig1}b, tensor-tensor and token-channel, can perform the efficient external scaling operation, but result in high quantization loss. The channel-channel quantization in Fig. \ref{fig1}c only causes minimal quantization loss ($< 0.1$) but lacks computational efficiency. We pre-execute the dequantization operation in the channel-channel quantization through weight updating to ensure computational efficiency, as shown in Fig. \ref{fig1}d. The perplexity of WikiText2 is measured with OPT-6.7B under INT8 quantization.}
\end{figure}

\subsection{Matrix multiplication after quantization}
We consider the matrix multiplication $WX$, where both activation $X\in R^{i\times n}$ and weight $W\in R^{n\times j}$ are subjected to the per-tensor symmetric quantization, as shown in Fig. \ref{fig1}a. The matrix multiplication after the per-tensor symmetric quantization can be expressed as 
$Y = (s^X \tilde X)(s^W \tilde W)$,
and $s^X, s^W$ are two constants that denote the scaling factors of the tensor $W$ and tensor $X$. Each element of output $Y$ can be calculated as: 
\begin{equation}\label{b}
\begin{aligned}
Y_{ij} = s^Xs^W \left(\sum\nolimits_{k = 1}^n {{\tilde X_{ik}}{ \tilde W_{kj}}}\right).
\end{aligned}
\end{equation}
In Eq. \eqref{b}, the scaling of dequantization  can be done efficiently outside the matrix multiplication via the scaling factors $s^X, s^W$ shared by the whole tensors. This approach ensures compatibility with the hardware-accelerated kernels, e.g. Cutlass Int8 GEMM, which only support external scaling within matrix multiplication. In the case of per-channel weight quantization and per-token activation quantization, as shown in Fig. \ref{fig1}b, which is currently the most widely studied quantization setting, the scaling factors $s^X\in R^i$ and $s^W\in R^j$ represent two vectors. Each element of output $Y$ can be calculated as,
\begin{equation}\label{c}
\begin{aligned}
Y_{ij} = s^X_is^W_j\left(\sum\nolimits_{k = 1}^n {{\tilde X_{ik}}{\tilde W_{kj}}}\right).
\end{aligned}
\end{equation}

Similarly, Eq. \eqref{c} allows external scaling of each output element via the weight scaling factor $s^W\in R^j$ shared per channel and the activation scaling factor $s^X\in R^i$ shared per token, and can be mapped to the hardware-accelerated kernels.

However, in the case of per-channel quantization of both weight and activation, as shown in Fig. \ref{fig1}c, with the scaling factors $s^X\in R^n$ and $s^W\in R^j$, each element of output $Y$ can be calculated as,
\begin{equation}\label{d}
\begin{aligned}
Y_{ij} = s^W_j \left(\sum\nolimits_{k = 1}^n  s^X_k{{\tilde X_{ik}}{\tilde W_{kj}}}\right).
\end{aligned}
\end{equation}
Here, the scaling factor  $s^X$  cannot be extracted for the external scaling of matrix multiplication, rendering it incompatible with hardware-acceleration kernels. In addition, we note that Eq. \eqref{d} introduces an additional multiplication operation, resulting in about $\{i\cdot j\cdot n\}$ times extra multiplication operations over the entire matrix multiplication.

\paragraph{Limitations  of different quantization dimensions.} 
\textbf{(i)} Although the per-tensor and per-token activation quantizations map the hardware-accelerated kernels in matrix multiplication well and perform efficient scaling, they lead to significant performance degradation of the quantized model. The intuitive reason is that the outlier channels dominate the quantization scales of tensor and token. For example, in the quantization examples shown in Fig. \ref{fig1}a and Fig. \ref{fig1}b, the outlier channel \{-60, -75, -93\} cause the numerical range of tensor or token to span widely, thereby  affecting  the rounding of normal values and adversely affecting the quantization. \textbf{(ii)} For the per-channel activation  quantization shown in Fig. \ref{fig1}c, the dominance of outlier channels is avoided by the efficient separation of the outlier channels. However, its internal scaling of matrix multiplication cannot be mapped to the hardware acceleration kernels (see Eq. \eqref{d}) and introduces additional computational overheads.

\section{Method}
We introduce OutlierTune, a novel framework for weight-activation quantization that addresses the challenges posed by the internal scaling of matrix multiplication, while preserving the accuracy of  per-channel activation quantization. Firstly, the pre-execution of dequantization  achieves the equivalent of internal scaling by tuning the corresponding weights, which facilitates the efficient implementation of per-channel activation quantization. Secondly, the symmetrization further reduces the discrepancy between the quantized results and the original  high-precision  expressions by balancing the numerical ranges of different outlier channels.

\subsection{Pre-execution of dequantization}\label{sec3.1}

\paragraph{Internal scaling based on Weight updates.} Considering the significant challenges arising from the internal scaling operation in the implementation of per-channel activation quantization, we find that it can be simplified by updating the associated weights. As shown in Fig. \ref{fig2}b, we avoid the need of internal scaling by updating the linear layer weights after Layernorm. The weight update process is conducted as follows:
\begin{equation}\label{e}
\begin{aligned}
W_{s} = W\odot  s^{X}
=W\odot\left[ {\begin{array}{*{20}{c}}
{{s_1}}&{{s_2}}& \cdots &{{s_n}}\\
{{s_1}}&{{s_2}}& \cdots &{{s_n}}\\
 \vdots & \vdots & \ddots & \vdots \\
{{s_1}}&{{s_2}}& \cdots &{{s_n}} 
\end{array}} \right],
\end{aligned}
\end{equation}
where $ s^{X} $ is a row vector representing the scaling factor associated with the per-channel activation quantization, $ \odot $ denotes  the element-wise multiplication. 
According to Eq. \eqref{e}, the scaling operations required for dequantization in Eq. \eqref{d} can be preemptively performed by integrating \( s^{X} \) into the weights of the linear layer following LayerNorm. This integration effectively embeds the scaling factor into the model parameters,  thereby streamlining the quantization process.
\paragraph{Matrix multiplication after per-channel quantization.} 
 Based on Eq. \eqref{e} and Eq. \eqref{d}, the matrix multiplication after the per-channel quantization can be writen as:
\begin{equation}\label{f}
\begin{aligned}
Y_{ij} = s^{W}_j \left(\sum\nolimits_{k = 1}^n  s^X_k{\left\lceil \frac{X_{ik}}{s^X_k}\right\rfloor  \left\lceil\frac{ W^T_{kj}}{s^{W}_j}\right\rfloor }\right)=s^{W_{s}}_j \left(\sum\nolimits_{k = 1}^n {\left\lceil \frac{X_{ik}}{s^X_k}\right\rfloor \left\lceil\frac{( W_{s}^T)_{kj}}{s^{W_{s}}_j}\right\rfloor}\right).
\end{aligned}
\end{equation}
Incorporating the weight updates \eqref{e} and matrix multiplication  \eqref{f}, we outline the quantization scheme for the per-channel activation quantization. This scheme involves updating weights through activation scaling factors while pre-performing the internal scaling operation. 
 By integrating the internal scaling process with the multiply-accumulate operation of matrix multiplication, we effectively eliminate the need for additional computation overheads associated with the per-channel activation quantization. It's noteworthy that both $ \tilde X_{ik}$ and $(\tilde W_s)_{kj}$ are represented by fixed-point representation and can be mapped to the efficient hardware-accelerated kernels to accelerate the matrix multiplication outlined in Eq. \eqref{f}.

\subsection{Symmetrization}\label{sec3.2}
As discussed in Section \ref{sec3.1}, Eq. \eqref{e} updates the weights via the activation scaling factors, effectively pre-executing the activation dequantization process. It's important to  note that the per-channel weight quantization occurs after the weight updates, which is inevitably affected by the difference in the activation scaling factors arising from the outlier channels and normal channels.

\begin{figure}[tbp]
\center{\includegraphics[width=13.95cm]  {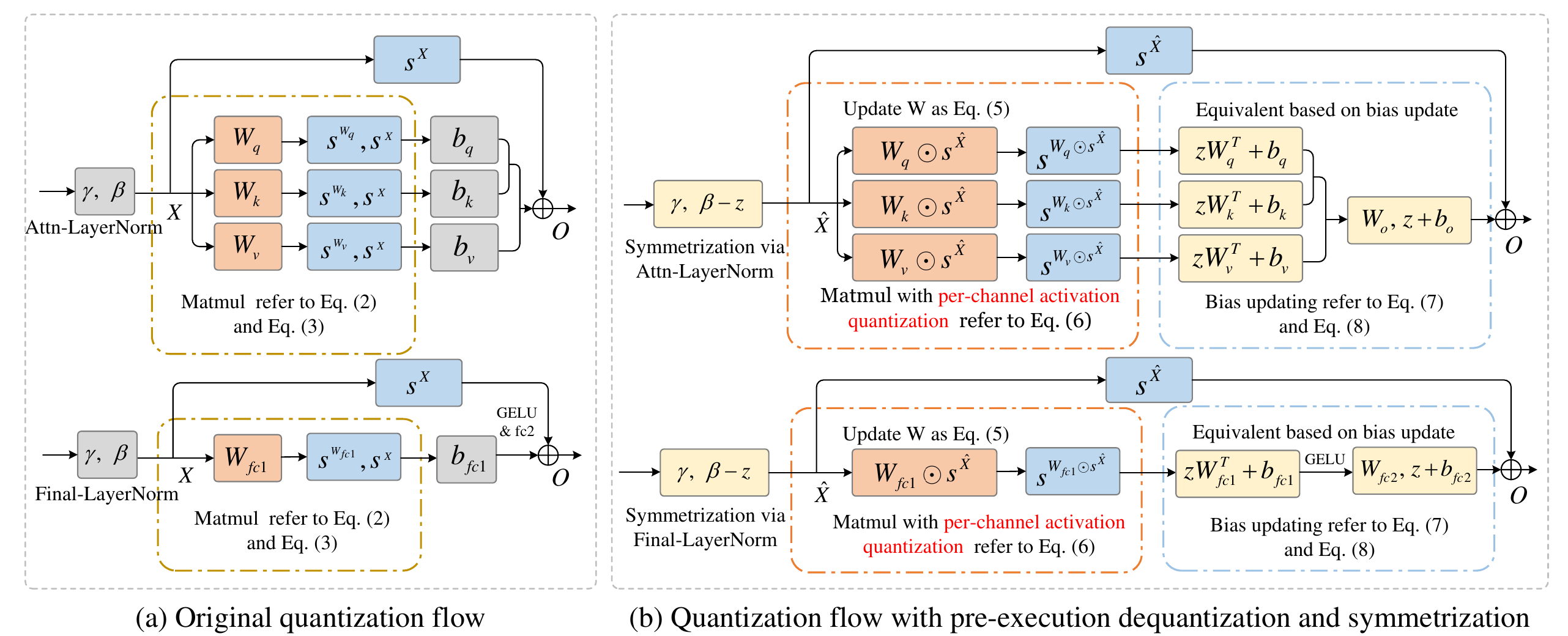}}
\caption{\label{fig2} Comparison of quantization flows of multi-head attention (MHA) and feed-forward network (FFN) under different quantization settings. OutlierTune avoids the additional computational overheads caused by the internal scaling required for per-channel quantization (right), thereby achieving similar efficiency as per-tensor or per-token quantization (left).}
\end{figure}

Considering the asymmetric numerical distribution of outlier channels, we use a symmetrization operation to equalize  scaling discrepancies between outlier and normal channels.
Inspired by \cite{wei-etal-2023-outlier}, we determine the symmetrization factors based on the maximum and minimum values of each activation channel, i.e. $z = (max(X_{:,k})+min(X_{:,k}))/2$, and combine it with the bias of LayerNorm to achieve the symmetrization of per-channel numerical range, i.e. $\hat X = X - z$. To maintain operational consistency, the adjustments are carried back to the network by updating the biases in the linear layers, ensuring that the changes in the front-end are mirrored in subsequent computations. The specific symmetrization process is shown in Fig. \ref{fig2}b. The linear mapping after symmetrization is as follows:
\begin{equation}\label{g}
\begin{aligned}
(\hat X+z)W^T + b \approx s^{W\odot s^{\hat X}}  \odot \left( \left\lceil \frac{\hat X}{s^{\hat X}}\right\rfloor  \left\lceil\frac{(W \odot s^{\hat X})^T}{s^{W\odot s^{\hat X}}}\right\rfloor\right) + (zW^T+b).
\end{aligned}
\end{equation}
When applying the residual connection after LayerNorm, we can achieve the equivalent by using the bias of $out$ and $fc2$ linear layers. The equivalence is shown as follows:
\begin{equation}\label{h}
\begin{aligned}
XW^T + b + (X_{res}^{sym}+z) \approx s^{W}s^{X}   \odot \left( \left\lceil \frac{X}{s^{X}}\right\rfloor  \left\lceil\frac{W^T}{s^{W}}\right\rfloor\right) + (b+z) +X_{res}^{sym},
\end{aligned}
\end{equation}
where $X_{res}^{sym}$ represents the residual connection after symmetrization. Since there are no obvious activation outliers in $out$ and $fc2$ linear layers, we use the quantization methods in Eq. \eqref{b} and Eq. \eqref{c} to perform the efficient matrix multiplication.

OutlierTune achieves efficient per-channel activation quantization by modifying the linear layers subsequent to LayerNorm, as detailed in Eq. \eqref{g}.  It's important to note that the per-channel activation quantizations $\left\lceil {\hat X}/{s^{\hat X}}\right\rfloor$ in Eq. \eqref{g} can be implemented offline through the kernel fusion \cite{wang2010kernel} with its previous LayerNorm module. Furthermore, the per-channel weight quantizations  $\left\lceil{(W_{l} \odot s^{\hat X})^T}/{s^{W_l\odot s^{\hat X}}}\right\rfloor$ can be predetermined offline once the static activation scaling factors are obtained. Therefore, our OutlierTune framework can achieve the same efficiency as per-tensor quantization while achieving low quantization loss of per-channel quantization.

\paragraph{Implementation.} 
OutlierTune is designed to optimize matrix multiplication following per-channel activation quantization, which is crucial for minimizing memory usage and enhancing model inference speed.
Given the presence of structured outliers exclusively in the output of LayerNorm,  we choose to use Eq. \eqref{g} for replacement of  the linear layers after LayerNorm. We employ W8A8 quantization for all matrix multiplication operations. In addition to lightweight operations such as ReLU, Softmax and LayerNorm, OutlierTune retains the biases of linear layers as FP16. We demonstrate the marginal impact of this choice on overall inference speed and memory usage in Sec. \ref{sec4.3}.

\section{Experiments}\label{sec4}
In this section, we present three sets of experiments aimed at validating the effectiveness of our OutlierTune framework. Sec. \ref{sec4.2} provides a comparative analysis between OutlierTune and existing methods across several zero-shot tasks and language datasets. In Sec. \ref{sec4.3}, we demonstrate the practical improvements in inference speed and memory usage through the implementation of the OutlierTune framework. Finally, Sec. \ref{sec4.4} explores the impact of each individual component.
\subsection{Experimental setup}\label{sec4.1}
\paragraph{Models and Datasets.}
This study rigorously evaluates the efficacy of our OutlierTune framework on two base models: Bloom \cite{workshop2022bloom}, OPT \cite{zhang2022opt}, and an instruction-tuned model: OPT-IML \cite{iyer2022opt}. Our evaluation  includes eight zero-shot benchmark tasks, including ARC (Challenge) \cite{clark2018think}, ARC (easy) \cite{clark2018think}, HellaSwag \cite{zellers2019hellaswag}, PIQA \cite{articlePiqa}, WinoGrande \cite{10.1145/3474381}, RTE \cite{wang-etal-2018-glue}, COPA \cite{roemmele2011choice}, BoolQ \cite{clark2019boolq}, and three language modeling datasets, WikiText2 \cite{DBLP:journals/corr/MerityXBS16}, Penn Treebank \cite{Marcus1994ThePT}, and C4 \cite{10.5555/3455716.3455856}. To ensure consistency and reliability in our evaluation process, we use the lm-eval-harness \cite{eval-harness} across all accuracy experiments.

\renewcommand{\arraystretch}{1.5}
\begin{table}[tbp]
\caption{\label{tab1}
We evaluate the perplexity of our OutlierTune framework on different OPT models. Our evaluation includes perplexity measurements across three different datasets: WikiText2, Ptb, and C4. These results are then compared  to two baselines to assess the performance improvements facilitated by our framework.}
\centering
\scalebox{0.59}{
\begin{tabular}{llllllllllllllllll}
\hline
\multicolumn{2}{c}{{Model}}  & \multicolumn{4}{c}{{OPT-6.7B}}  & \multicolumn{4}{c}{{OPT-13B}}& \multicolumn{4}{c}{{OPT-30B}} &  \multicolumn{4}{c}{{OPT-66B}} \\

\cmidrule(lr){1-2}\cmidrule(lr){3-6}\cmidrule(lr){7-10}\cmidrule(lr){11-14}\cmidrule(lr){15-18}

 \multicolumn{2}{c}{{Task/PPL$\downarrow$}} & \makecell[c]{{FP16}} & \makecell[c]{{Int8*}} & \makecell[c]{{Int8}} & \makecell[c]{{Int6}} & \makecell[c]{{FP16}} & \makecell[c]{{Int8*}} & \makecell[c]{{Int8}} & \makecell[c]{{Int6}} & \makecell[c]{{FP16}} & \makecell[c]{{Int8*}} & \makecell[c]{{Int8}} & \makecell[c]{{Int6}} & \makecell[c]{{FP16}} & \makecell[c]{{Int8*}} & \makecell[c]{{Int8}} & \makecell[c]{{Int6}}\\
\hline
 \multirow{3}{*}{Wiki2} & \makecell[c]{LLM.int8}&  \multirow{3}{*}{10.86} & \makecell[c]{\textcolor{gray}{-}} & \makecell[c]{\textcolor{gray}{10.86}} & \makecell[c]{\textcolor{gray}{-}}& \multirow{3}{*}{10.13} & \makecell[c]{\textcolor{gray}{-}} & \makecell[c]{\textcolor{gray}{10.13}} & \makecell[c]{\textcolor{gray}{-}}&  \multirow{3}{*}{9.56}  &\makecell[c]{\textcolor{gray}{-}} & \makecell[c]{\textcolor{gray}{9.57}} & \makecell[c]{\textcolor{gray}{-}} &  \multirow{3}{*}{9.34} & \makecell[c]{\textcolor{gray}{-}} & \makecell[c]{\textcolor{gray}{9.35}} & \makecell[c]{\textcolor{gray}{-}}\\

                       & \makecell[c]{Smoothquant}&         & \makecell[c]{{10.89}} & \makecell[c]{{10.88}} & \makecell[c]{12.55}&        & \makecell[c]{10.37} & \makecell[c]{10.37} & \makecell[c]{{13.60}}&         &\makecell[c]{9.59}&\makecell[c]{9.59} & \makecell[c]{127.57} &       & \makecell[c]{{9.80}}& \makecell[c]{{9.79}} & \makecell[c]{{1002.23}}\\

                            & \makecell[c]{Ours}&    &  \makecell[c]{\textbf{10.87}} & \makecell[c]{\textbf{10.87}} &\makecell[c]{\textbf{11.49}}&                 & \makecell[c]{\textbf{10.18}} & \makecell[c]{\textbf{10.14}} & \makecell[c]{\textbf{11.78}} &                 &\makecell[c]{\textbf{9.59}} & \makecell[c]{\textbf{9.57}} & \makecell[c]{\textbf{10.26}} &              & \makecell[c]{\textbf{9.35}}& \makecell[c]{\textbf{9.34}} & \makecell[c]{\textbf{9.67}}\\

\hline
 \multirow{3}{*}{Ptb} & \makecell[c]{LLM.int8}&  \multirow{3}{*}{13.09} & \makecell[c]{\textcolor{gray}{-}} & \makecell[c]{\textcolor{gray}{13.12}} & \makecell[c]{\textcolor{gray}{-}}& \multirow{3}{*}{12.34} & \makecell[c]{\textcolor{gray}{-}} & \makecell[c]{\textcolor{gray}{12.35}} & \makecell[c]{\textcolor{gray}{-}} &  \multirow{3}{*}{11.84}  &\makecell[c]{\textcolor{gray}{-}} & \makecell[c]{\textcolor{gray}{11.85}} & \makecell[c]{\textcolor{gray}{-}} &  \multirow{3}{*}{11.36} & \makecell[c]{\textcolor{gray}{-}} & \makecell[c]{\textcolor{gray}{11.37}} & \makecell[c]{\textcolor{gray}{-}}\\

                     & \makecell[c]{Smoothquant}&       &  \makecell[c]{{13.14}} & \makecell[c]{{13.14}} & \makecell[c]{16.10}&      & \makecell[c]{12.57} & \makecell[c]{12.57} & \makecell[c]{{23.04}}&         &\makecell[c]{11.88}&\makecell[c]{11.88} & \makecell[c]{215.45} &       & \makecell[c]{{11.67}}& \makecell[c]{{11.67}} & \makecell[c]{{707.30}}\\

                            & \makecell[c]{Ours}&    &   \makecell[c]{\textbf{13.10}} & \makecell[c]{\textbf{13.10}} &\makecell[c]{\textbf{13.83}}&                 & \makecell[c]{\textbf{12.36}} & \makecell[c]{\textbf{12.34}} & \makecell[c]{\textbf{16.29}} &                 &\makecell[c]{\textbf{11.85}} & \makecell[c]{\textbf{11.85}} & \makecell[c]{\textbf{13.30}} &              & \makecell[c]{\textbf{11.36}}& \makecell[c]{\textbf{11.37}} & \makecell[c]{\textbf{11.52}}\\

\hline
 \multirow{3}{*}{C4}  & \makecell[c]{LLM.int8}&  \multirow{3}{*}{11.74} & \makecell[c]{\textcolor{gray}{-}} & \makecell[c]{\textcolor{gray}{11.75}} & \makecell[c]{\textcolor{gray}{-}} & \multirow{3}{*}{11.20} & \makecell[c]{\textcolor{gray}{-}} & \makecell[c]{\textcolor{gray}{11.21}} & \makecell[c]{\textcolor{gray}{-}}  &  \multirow{3}{*}{10.69}  &\makecell[c]{\textcolor{gray}{-}} & \makecell[c]{\textcolor{gray}{10.70}} & \makecell[c]{\textcolor{gray}{-}} &  \multirow{3}{*}{10.28} & \makecell[c]{\textcolor{gray}{-}} & \makecell[c]{\textcolor{gray}{10.29}} & \makecell[c]{\textcolor{gray}{-}}\\

              & \makecell[c]{Smoothquant} &          & \makecell[c]{{11.80}} & \makecell[c]{{11.80}} & \makecell[c]{14.16}&           & \makecell[c]{11.24} & \makecell[c]{11.24} & \makecell[c]{{18.33}}&           &\makecell[c]{10.72}&\makecell[c]{10.72} & \makecell[c]{255.23} &            & \makecell[c]{{10.47}}& \makecell[c]{{10.47}} & \makecell[c]{{844.24}}\\

                           & \makecell[c]{Ours} &    &  \makecell[c]{\textbf{11.75}} & \makecell[c]{\textbf{11.75}} &\makecell[c]{\textbf{12.22}}&                 & \makecell[c]{\textbf{11.21}} & \makecell[c]{\textbf{11.21}} & \makecell[c]{\textbf{12.43}} &                 &\makecell[c]{\textbf{10.70}} & \makecell[c]{\textbf{10.70}} & \makecell[c]{\textbf{11.13}} &              & \makecell[c]{\textbf{10.29}}& \makecell[c]{\textbf{10.29}} & \makecell[c]{\textbf{10.45}}\\

\hline
\end{tabular}}
\end{table}

\begin{table}[tbp]
\caption{\label{tab2}
The accuracy of our OutlierTune framework is compared with the baselines on 8 different zero-shot tasks. Int8* represents the weight of per-channel symmetric quantization. Since LLM.int8 involves mixed-precision dynamic quantization, we only give the results without comparison.}
\centering
\scalebox{0.58}{
\begin{tabular}{p{1.6cm}lllllllllllllllll}
\hline
\multicolumn{2}{c}{{Model}}  & \multicolumn{4}{c}{{OPT-6.7B}}  & \multicolumn{4}{c}{{OPT-13B}}& \multicolumn{4}{c}{{OPT-30B}} &  \multicolumn{4}{c}{{OPT-66B}} \\

\cmidrule(lr){1-2}\cmidrule(lr){3-6}\cmidrule(lr){7-10}\cmidrule(lr){11-14}\cmidrule(lr){15-18}

 \multicolumn{1}{c}{{Task/Acc$\uparrow$}} & \multicolumn{1}{c}{{Methods}} & \makecell[c]{{FP16}} & \makecell[c]{{Int8*}} & \makecell[c]{{Int8}} & \makecell[c]{{Int6}} & \makecell[c]{{FP16}} & \makecell[c]{{Int8*}} & \makecell[c]{{Int8}} & \makecell[c]{{Int6}} & \makecell[c]{{FP16}} & \makecell[c]{{Int8*}} & \makecell[c]{{Int8}} & \makecell[c]{{Int6}} & \makecell[c]{{FP16}} & \makecell[c]{{Int8*}} & \makecell[c]{{Int8}} & \makecell[c]{{Int6}}\\

\hline
 \multirow{3}{*}{\makecell*[l]{ARC\\ (Challenge)}} & \makecell[c]{LLM.int8}&  \multirow{3}{*}{30.55} & \makecell[c]{\textcolor{gray}{-}} & \makecell[c]{\textcolor{gray}{30.80}} & \makecell[c]{\textcolor{gray}{-}} & \multirow{3}{*}{33.02} & \makecell[c]{\textcolor{gray}{-}} & \makecell[c]{\textcolor{gray}{33.11}} & \makecell[c]{\textcolor{gray}{-}}  &  \multirow{3}{*}{34.73}  &\makecell[c]{\textcolor{gray}{-}} & \makecell[c]{\textcolor{gray}{35.07}} & \makecell[c]{\textcolor{gray}{-}} &  \multirow{3}{*}{37.29} & \makecell[c]{\textcolor{gray}{-}} & \makecell[c]{\textcolor{gray}{37.63}} & \makecell[c]{\textcolor{gray}{-}}\\

                       & \makecell[c]{Smoothquant}&         & \makecell[c]{\textbf{30.20}} & \makecell[c]{{30.20}} & \makecell[c]{30.12}&        & \makecell[c]{32.51} & \makecell[c]{32.25} & \makecell[c]{{25.60}}&         &\makecell[c]{\textbf{34.22}}&\makecell[c]{33.62} & \makecell[c]{19.62} &       & \makecell[c]{\textbf{37.12}}& \makecell[c]{\textbf{37.12}} & \makecell[c]{{20.82}}\\

                            & \makecell[c]{Ours}&    &  \makecell[c]{{30.12}} & \makecell[c]{\textbf{30.97}} &\makecell[c]{\textbf{31.06}}&                 & \makecell[c]{\textbf{33.45}} & \makecell[c]{\textbf{33.70}} & \makecell[c]{\textbf{32.25}} &                 &\makecell[c]{{33.96}} & \makecell[c]{\textbf{34.47}} & \makecell[c]{\textbf{32.94}} &              & \makecell[c]{{37.03}}& \makecell[c]{{36.77}} & \makecell[c]{\textbf{36.77}}\\

\hline
 \multirow{3}{*}{\makecell*[c]{ARC\\ (Easy)}} & \makecell[c]{LLM.int8}&  \multirow{3}{*}{65.66} & \makecell[c]{\textcolor{gray}{-}} & \makecell[c]{\textcolor{gray}{65.74}} & \makecell[c]{\textcolor{gray}{-}} & \multirow{3}{*}{67.05} & \makecell[c]{\textcolor{gray}{-}} & \makecell[c]{\textcolor{gray}{66.96}} & \makecell[c]{\textcolor{gray}{-}}  &  \multirow{3}{*}{70.08}  &\makecell[c]{\textcolor{gray}{-}} & \makecell[c]{\textcolor{gray}{69.82}} & \makecell[c]{\textcolor{gray}{-}} &  \multirow{3}{*}{71.63} & \makecell[c]{\textcolor{gray}{-}} & \makecell[c]{\textcolor{gray}{71.46}} & \makecell[c]{\textcolor{gray}{-}}\\

                     & \makecell[c]{Smoothquant}&       &  \makecell[c]{{65.49}} & \makecell[c]{\textbf{65.49}} & \makecell[c]{63.80}&      & \makecell[c]{66.54} & \makecell[c]{66.58} & \makecell[c]{{51.35}}&         &\makecell[c]{69.57}&\makecell[c]{\textbf{69.82}} & \makecell[c]{40.95} &       & \makecell[c]{{70.88}}& \makecell[c]{{71.00}} & \makecell[c]{{27.69}}\\

                            & \makecell[c]{Ours}&    &   \makecell[c]{\textbf{65.87}} & \makecell[c]{{65.45}} &\makecell[c]{\textbf{65.45}}&                 & \makecell[c]{\textbf{67.21}} & \makecell[c]{\textbf{67.21}} & \makecell[c]{\textbf{66.16}} &                 &\makecell[c]{\textbf{69.78}} & \makecell[c]{{69.74}} & \makecell[c]{\textbf{67.97}} &              & \makecell[c]{\textbf{72.01}}& \makecell[c]{\textbf{71.72}} & \makecell[c]{\textbf{71.13}}\\

\hline
 \multirow{3}{*}{BoolQ}  & \makecell[c]{LLM.int8}&  \multirow{3}{*}{65.99} & \makecell[c]{\textcolor{gray}{-}} & \makecell[c]{\textcolor{gray}{65.78}} & \makecell[c]{\textcolor{gray}{-}} & \multirow{3}{*}{65.81} & \makecell[c]{\textcolor{gray}{-}} & \makecell[c]{\textcolor{gray}{65.96}} & \makecell[c]{\textcolor{gray}{-}}  &  \multirow{3}{*}{70.37}  &\makecell[c]{\textcolor{gray}{-}} & \makecell[c]{\textcolor{gray}{70.33}} & \makecell[c]{\textcolor{gray}{-}} &  \multirow{3}{*}{69.66} & \makecell[c]{\textcolor{gray}{-}} & \makecell[c]{\textcolor{gray}{69.54}} & \makecell[c]{\textcolor{gray}{-}}\\

              & \makecell[c]{Smoothquant} &          & \makecell[c]{\textbf{66.18}} & \makecell[c]{\textbf{66.36}} & \makecell[c]{62.63}&           & \makecell[c]{64.95} & \makecell[c]{65.11} & \makecell[c]{\textbf{66.97}}&           &\makecell[c]{\textbf{70.83}}&\makecell[c]{\textbf{71.01}} & \makecell[c]{50.55} &            & \makecell[c]{{68.20}}& \makecell[c]{{68.07}} & \makecell[c]{{42.66}}\\

                           & \makecell[c]{Ours} &    &  \makecell[c]{{65.93}} & \makecell[c]{{66.12}} &\makecell[c]{\textbf{62.72}}&                 & \makecell[c]{\textbf{65.81}} & \makecell[c]{\textbf{66.12}} & \makecell[c]{{61.34}} &                 &\makecell[c]{{70.55}} & \makecell[c]{{70.12}} & \makecell[c]{\textbf{68.84}} &              & \makecell[c]{\textbf{69.69}}& \makecell[c]{\textbf{69.63}} & \makecell[c]{\textbf{69.42}}\\

\hline
 \multirow{3}{*}{\makecell[c]{Copa}}  & \makecell[c]{LLM.int8}&  \multirow{3}{*}{81.00} & \makecell[c]{\textcolor{gray}{-}} & \makecell[c]{\textcolor{gray}{81.00}} & \makecell[c]{\textcolor{gray}{-}} & \multirow{3}{*}{86.00} & \makecell[c]{\textcolor{gray}{-}} & \makecell[c]{\textcolor{gray}{86.00}} & \makecell[c]{\textcolor{gray}{-}}  &  \multirow{3}{*}{82.00}  &\makecell[c]{\textcolor{gray}{-}} & \makecell[c]{\textcolor{gray}{82.00}} & \makecell[c]{\textcolor{gray}{-}} &  \multirow{3}{*}{86.00} & \makecell[c]{\textcolor{gray}{-}} & \makecell[c]{\textcolor{gray}{86.00}} & \makecell[c]{\textcolor{gray}{-}}\\

              & \makecell[c]{Smoothquant} &          & \makecell[c]{{82.00}} & \makecell[c]{\textbf{83.00}} & \makecell[c]{\textbf{85.00}}&           & \makecell[c]{85.00} & \makecell[c]{85.00} & \makecell[c]{{64.00}}&           &\makecell[c]{82.00}&\makecell[c]{\textbf{84.00}} & \makecell[c]{65.00} &            & \makecell[c]{{84.00}}& \makecell[c]{{84.00}} & \makecell[c]{{54.00}}\\

                           & \makecell[c]{Ours} &    &  \makecell[c]{\textbf{82.00}} & \makecell[c]{{81.00}} &\makecell[c]{{83.00}}&                 & \makecell[c]{\textbf{86.00}} & \makecell[c]{\textbf{86.00}} & \makecell[c]{\textbf{85.00}} &                 &\makecell[c]{\textbf{82.00}} & \makecell[c]{{83.00}} & \makecell[c]{\textbf{84.00}} &              & \makecell[c]{\textbf{85.00}}& \makecell[c]{\textbf{86.00}} & \makecell[c]{\textbf{85.00}}\\

\hline
 \multirow{3}{*}{HellaSwag}  & \makecell[c]{LLM.int8}&  \multirow{3}{*}{50.47} & \makecell[c]{\textcolor{gray}{-}} & \makecell[c]{\textcolor{gray}{50.48}} & \makecell[c]{\textcolor{gray}{-}} & \multirow{3}{*}{52.46} & \makecell[c]{\textcolor{gray}{-}} & \makecell[c]{\textcolor{gray}{52.44}} & \makecell[c]{\textcolor{gray}{-}}  &  \multirow{3}{*}{54.28}  &\makecell[c]{\textcolor{gray}{-}} & \makecell[c]{\textcolor{gray}{54.32}} & \makecell[c]{\textcolor{gray}{-}} &  \multirow{3}{*}{56.42} & \makecell[c]{\textcolor{gray}{-}} & \makecell[c]{\textcolor{gray}{56.38}} & \makecell[c]{\textcolor{gray}{-}}\\

              & \makecell[c]{Smoothquant} &          & \makecell[c]{{50.44}} & \makecell[c]{{50.39}} & \makecell[c]{48.19}&           & \makecell[c]{52.15} & \makecell[c]{52.20} & \makecell[c]{{45.17}}&           &\makecell[c]{54.18}&\makecell[c]{54.30} & \makecell[c]{28.05} &            & \makecell[c]{\textbf{56.39}}& \makecell[c]{{56.23}} & \makecell[c]{{26.58}}\\

                           & \makecell[c]{Ours} &    &  \makecell[c]{\textbf{50.50}} & \makecell[c]{\textbf{50.55}} &\makecell[c]{\textbf{49.90}}&                 & \makecell[c]{\textbf{52.28}} & \makecell[c]{\textbf{52.37}} & \makecell[c]{\textbf{48.90}} &                 &\makecell[c]{\textbf{54.35}} & \makecell[c]{\textbf{54.33}} & \makecell[c]{\textbf{53.29}} &              & \makecell[c]{{56.31}}& \makecell[c]{\textbf{56.38}} & \makecell[c]{\textbf{55.69}}\\

\hline
 \multirow{3}{*}{PIQA}  &  \makecell[c]{LLM.int8}&  \multirow{3}{*}{76.22} & \makecell[c]{\textcolor{gray}{-}} & \makecell[c]{\textcolor{gray}{76.28}} & \makecell[c]{\textcolor{gray}{-}} & \multirow{3}{*}{76.01} & \makecell[c]{\textcolor{gray}{-}} & \makecell[c]{\textcolor{gray}{75.90}} & \makecell[c]{\textcolor{gray}{-}}  &  \multirow{3}{*}{77.58}  &\makecell[c]{\textcolor{gray}{-}} & \makecell[c]{\textcolor{gray}{77.64}} & \makecell[c]{\textcolor{gray}{-}} &  \multirow{3}{*}{78.73} & \makecell[c]{\textcolor{gray}{-}} & \makecell[c]{\textcolor{gray}{78.62}} & \makecell[c]{\textcolor{gray}{-}}\\

              & \makecell[c]{Smoothquant} &          & \makecell[c]{\textbf{76.28}} & \makecell[c]{{75.95}} & \makecell[c]{74.97}&           & \makecell[c]{75.68} & \makecell[c]{75.57} & \makecell[c]{{67.41}}&           &\makecell[c]{77.48}&\makecell[c]{\textbf{77.69}} & \makecell[c]{57.51} &            & \makecell[c]{{78.29}}& \makecell[c]{{78.84}} & \makecell[c]{{53.26}}\\

                           & \makecell[c]{Ours} &    &  \makecell[c]{{76.22}} & \makecell[c]{\textbf{76.39}} &\makecell[c]{\textbf{76.12}}&                 & \makecell[c]{\textbf{76.27}} & \makecell[c]{\textbf{75.63}} & \makecell[c]{\textbf{75.24}} &                 &\makecell[c]{\textbf{77.53}} & \makecell[c]{{77.31}} & \makecell[c]{\textbf{76.77}} &              & \makecell[c]{\textbf{78.94}}& \makecell[c]{\textbf{78.89}} & \makecell[c]{\textbf{78.24}}\\

\hline
 \multirow{3}{*}{RTE}  &  \makecell[c]{LLM.int8}&  \multirow{3}{*}{55.23} & \makecell[c]{\textcolor{gray}{-}} & \makecell[c]{\textcolor{gray}{55.60}} & \makecell[c]{\textcolor{gray}{-}} & \multirow{3}{*}{58.12} & \makecell[c]{\textcolor{gray}{-}} & \makecell[c]{\textcolor{gray}{58.12}} & \makecell[c]{\textcolor{gray}{-}}  &  \multirow{3}{*}{57.76}  &\makecell[c]{\textcolor{gray}{-}} & \makecell[c]{\textcolor{gray}{59.21}} & \makecell[c]{\textcolor{gray}{-}} &  \multirow{3}{*}{60.29} & \makecell[c]{\textcolor{gray}{-}} & \makecell[c]{\textcolor{gray}{61.01}} & \makecell[c]{\textcolor{gray}{-}}\\

              & \makecell[c]{Smoothquant} &          & \makecell[c]{{55.23}} & \makecell[c]{{55.60}} & \makecell[c]{53.43}&           & \makecell[c]{54.87} & \makecell[c]{57.04} & \makecell[c]{{53.79}}&           &\makecell[c]{58.12}&\makecell[c]{56.32} & \makecell[c]{52.34} &            & \makecell[c]{\textbf{62.45}}& \makecell[c]{\textbf{61.73}} & \makecell[c]{{52.35}}\\

                           & \makecell[c]{Ours} &    &  \makecell[c]{\textbf{56.32}} & \makecell[c]{\textbf{56.32}} &\makecell[c]{\textbf{54.51}}&                 & \makecell[c]{\textbf{58.12}} & \makecell[c]{\textbf{58.12}} & \makecell[c]{\textbf{55.23}} &                 &\makecell[c]{\textbf{60.29}} & \makecell[c]{\textbf{59.21}} & \makecell[c]{\textbf{56.68}} &              & \makecell[c]{{60.64}}& \makecell[c]{{61.37}} & \makecell[c]{\textbf{59.21}}\\

\hline
 \multirow{3}{*}{WinoGrande}  & \makecell[c]{LLM.int8}&  \multirow{3}{*}{65.43} & \makecell[c]{\textcolor{gray}{-}} & \makecell[c]{\textcolor{gray}{64.48}} & \makecell[c]{\textcolor{gray}{-}} & \multirow{3}{*}{65.19} & \makecell[c]{\textcolor{gray}{-}} & \makecell[c]{\textcolor{gray}{65.11}} & \makecell[c]{\textcolor{gray}{-}}  &  \multirow{3}{*}{68.35}  &\makecell[c]{\textcolor{gray}{-}} & \makecell[c]{\textcolor{gray}{67.80}} & \makecell[c]{\textcolor{gray}{-}} &  \multirow{3}{*}{68.82} & \makecell[c]{\textcolor{gray}{-}} & \makecell[c]{\textcolor{gray}{68.82}} & \makecell[c]{\textcolor{gray}{-}}\\

              & \makecell[c]{Smoothquant} &          & \makecell[c]{\textbf{66.14}} & \makecell[c]{\textbf{66.06}} & \makecell[c]{\textbf{66.38}}&           & \makecell[c]{\textbf{65.35}} & \makecell[c]{\textbf{65.04}} & \makecell[c]{{54.38}}&           &\makecell[c]{68.11}&\makecell[c]{\textbf{68.67}} & \makecell[c]{51.62} &            & \makecell[c]{{67.64}}& \makecell[c]{{68.03}} & \makecell[c]{{48.86}}\\

                           & \makecell[c]{Ours} &    &  \makecell[c]{{64.64}} & \makecell[c]{{64.72}} &\makecell[c]{{63.06}}&                 & \makecell[c]{{64.33}} & \makecell[c]{{64.96}} & \makecell[c]{\textbf{64.72}} &                 &\makecell[c]{\textbf{68.35}} & \makecell[c]{{68.35}} & \makecell[c]{\textbf{65.43}} &              & \makecell[c]{\textbf{68.82}}& \makecell[c]{\textbf{69.14}} & \makecell[c]{\textbf{67.72}}\\

\hline
\end{tabular}}
\end{table}

\paragraph{Baseline.} 
Our primary baselines include LLM.int8 \cite{dettmers2022llmint8} and SmoothQuant \citep{pmlr-v202-xiao23c}. LLM.int8 adopts the per-token activation quantization while retaining the outlier channels as FP16 format. This approach achieves favorable results in Int8 quantization by isolating the  effect  of outlier channels, but its delay performance is  worse due to the mixed-precision effect. SmoothQuant migrates the quantization difficulty from activations to weights by a mathematically equivalent transformation, thereby achieving near-lossless int8 quantization. Other methods, such as RTN and ZeroQuant \cite{NEURIPS2022_adf7fa39}, have not shown superior performance compared to SmoothQuant and were therefore  excluded from our analysis.

\paragraph{Quantization setting.} 
Unless explicitly specified, OutlierTune employs the per-channel symmetric quantization for activations after LayerNorm and the per-token symmetric quantization for other activations. This decision is based on the observation that the  activation outliers  occur predominantly after LayerNorm \citep{wei2022outlier}. For the baselines, all activations utilize the per-token symmetric quantization method. In order to optimise latency and accuracy, we explore two different weight quantization strategies. The first, denoted by Int8*, adopts the per-channel symmetric quantization, prioritizing inference speed. The second, denoted by Int8, employs the per-channel asymmetric quantization, emphasizing high performance. More implementation details can be found in Appendix \ref{A}.

\subsection{Accuracy Evaluation}\label{sec4.2}
\paragraph{Results of OPTs.}
We investigated the effect of OutlierTune by compressing the weights and activations of OPT models into 6 and 8 bits. Subsequently, we evaluated the model's perplexity across three language tasks (see Table \ref{tab1}), and its accuracy on 8 zero-shot tasks (see Table \ref{tab2}). For Table \ref{tab1}, our OutlierTune framework slightly outperforms the Smoothquant and matches the performance of LLM.int8 in Int8 quantization. It's noteworthy that, given the negative impact of LLM.int8 on inference latency in most cases, we only provide results for LLM.int8 without a direct comparison.
 Smoothquant shows a notable performance degradation when quantizing to int6, especially for the larger models. This degradation is likely attributable to the presence of more significant outliers in larger models,  making the accurate quantization more difficult. Interestingly, our OutlierTune framework appears to be more adaptable to the larger models. For example, OutlierTune results in a minimal perplexity increase of only 0.16 on the OPT-66B model under Int6 quantization.

In Table \ref{tab2}, OutlierTune shows accuracy levels comparable to LLM.int8 in int8 quantization and outperforms Smoothquant in most tasks. Despite the challenges encountered with Smoothquant in int6 quantization, OutlierTune can still maintain the accuracy of different tasks, with significant improvements observed: 43.44\% in ARC (Easy), 31\% in Copa, and 26.76\% in BoolQ.  In summary, our proposed OutlierTune framework achieves performance close to FP16 in Int8 quantization and only causes about 1-2 points of accuracy loss in Int6 quantization.

\begin{table}[tbp]
\caption{\label{tab3}
The performance of our OutlierTune framework on Bloom-7B1.}
\centering
\scalebox{0.73}{
\begin{tabular}{llllllllllllllllll}
\hline
 \multicolumn{1}{c}{{Task/Acc$\uparrow$}}  & \makecell[c]{{ARC(Challenge)}} & \makecell[c]{{ARC(Easy)}} & \makecell[c]{{BoolQ}} & \makecell[c]{{Copa}} & \makecell[c]{{HellaSwag}} & \makecell[c]{{PIQA}} & \makecell[c]{{RTE}} & \makecell[c]{{WinoGrande}} \\
\hline

 \makecell[c]{FP16} &  \makecell[c]{{30.29}} & \makecell[c]{{65.03}} &\makecell[c]{{62.91}}&  \makecell[c]{{72.00}} &  \makecell[c]{{46.27}} & \makecell[c]{{72.63}} & \makecell[c]{{54.15}} & \makecell[c]{{64.56}}\\

\hline
 \makecell[c]{Smoothquant-Int8} &  \makecell[c]{\textbf{30.38}} & \makecell[c]{{64.86}} & \makecell[c]{\textbf{63.15}} &  \makecell[c]{71.00}  &  \makecell[c]{\textbf{46.18}}  & \makecell[c]{\textbf{73.07}} & \makecell[c]{54.16} & \makecell[c]{{63.30}}\\

 \makecell[c]{Ours-Int8} & \makecell[c]{{29.52}} & \makecell[c]{\textbf{64.98}} &\makecell[c]{{63.12}} &  \makecell[c]{\textbf{71.00}}  & \makecell[c]{{46.08}} & \makecell[c]{{72.52}} & \makecell[c]{\textbf{54.51}} & \makecell[c]{\textbf{63.69}}\\

\hline
 \makecell[c]{Smoothquant-Int6} &  \makecell[c]{{29.43}} & \makecell[c]{{61.91}} & \makecell[c]{61.44} &  \makecell[c]{56.00}  &  \makecell[c]{44.57}  & \makecell[c]{72.03} & \makecell[c]{53.79} & \makecell[c]{{59.51}} &  \\

\makecell[c]{Ours-Int6} & \makecell[c]{\textbf{29.61}} & \makecell[c]{\textbf{63.09}} &\makecell[c]{\textbf{62.63}} &  \makecell[c]{\textbf{58.00}}  & \makecell[c]{\textbf{45.14}} & \makecell[c]{\textbf{72.31}} & \makecell[c]{\textbf{55.60}} & \makecell[c]{\textbf{61.23}} \\
\hline
\end{tabular}}
\end{table}

\paragraph{Results of different LLMs.} 
To illustrate the generalizability of our OutlierTune framework across different models, we show  the evaluation results on eight zero-shot tasks  by quantizing Bloom-7B1 to 6 and 8 bits, as shown in Table \ref{tab3}. The quantization challenge posed by Bloom-7B1 is comparable to that of OPT-6.7B: both the baseline and our OutlierTune framework can guarantee an average accuracy loss of less than 0.3\% under int8 quantization compared to FP16. Upon further quantized to int6, OutlierTune demonstrates superior robustness, consistently outperforming Smoothquant across all tasks, with an average accuracy improvement of 2\%.  These results highlight  the adaptability of the OutlierTune framework across different model architectures.

\begin{table}[tbp]
\caption{\label{tab4}
The performance of our OutlierTune framework on OPT-IML-30B.}
\centering
\scalebox{0.75}{
\begin{tabular}{llllllllllllllllll}
\hline
 \multicolumn{1}{c}{{Task/Acc$\uparrow$}}  & \makecell[c]{{ARC(Challenge)}} & \makecell[c]{{ARC(Easy)}} & \makecell[c]{{BoolQ}} & \makecell[c]{{Copa}} & \makecell[c]{{HellaSwag}} & \makecell[c]{{PIQA}} & \makecell[c]{{RTE}} & \makecell[c]{{WinoGrande}}\\
\hline

 \makecell[c]{FP16} &  \makecell[c]{{38.31}} & \makecell[c]{{71.46}} &\makecell[c]{{79.82}}&  \makecell[c]{{79.00}} &  \makecell[c]{{52.70}} & \makecell[c]{{76.93}} & \makecell[c]{{81.95}} & \makecell[c]{{67.80}} \\

\hline
 \makecell[c]{Smoothquant-Int8} &  \makecell[c]{\textbf{38.05}} & \makecell[c]{\textbf{71.76}} & \makecell[c]{{79.48}} &  \makecell[c]{79.00}  &  \makecell[c]{\textbf{52.84}}  & \makecell[c]{\textbf{77.04}} & \makecell[c]{79.42} & \makecell[c]{\textbf{69.06}} \\

 \makecell[c]{Ours-Int8} & \makecell[c]{{37.80}} & \makecell[c]{{71.30}} &\makecell[c]{\textbf{80.03}} &  \makecell[c]{\textbf{79.00}}  & \makecell[c]{{52.72}} & \makecell[c]{{76.66}} & \makecell[c]{\textbf{80.87}} & \makecell[c]{{68.43}}\\

\hline
 \makecell[c]{Smoothquant-Int6} &  \makecell[c]{{30.46}} & \makecell[c]{{60.94}} & \makecell[c]{70.73} &  \makecell[c]{75.00}  &  \makecell[c]{39.55}  & \makecell[c]{70.18} & \makecell[c]{70.75} & \makecell[c]{{59.66}} \\

 \makecell[c]{Ours-Int6} & \makecell[c]{\textbf{38.40}} & \makecell[c]{\textbf{71.51}} & \makecell[c]{\textbf{79.27}} &  \makecell[c]{\textbf{80.00}}  &  \makecell[c]{\textbf{52.41}}  & \makecell[c]{\textbf{76.06}} & \makecell[c]{\textbf{80.87}} & \makecell[c]{\textbf{68.19}}\\
\hline
\end{tabular}}
\end{table}

\paragraph{Results of Instruction-Tuned LLM.} 
We now investigate the effectiveness of OutlierTune on the instruction-tuned model, OPT-IML-30B \cite{iyer2022opt}, as detailed in Table \ref{tab4}.  The results show that both OutlierTune and the baseline method preserve model accuracy under int8 quantization, with the average accuracy loss across various tasks remaining below 0.2\%.  Notably, OutlierTune continues to preserve the model accuracy under more restrictive INT6 quantization, limiting the average accuracy loss to just 0.16\%, whereas the baseline demonstrates a significant reduction, with an average accuracy loss of approximately 10\%. We speculate that this may be attributed to the training of instruction-tuned LLM, which reduces the numerical range of outliers, thus facilitating easier quantification of the model. The results in Table \ref{tab4} underline the suitability of OutlierTune,  highlighting its potential to minimize the accuracy losses during aggressive quantizations.

\subsection{Speedup and Memory Saving}\label{sec4.3}
\begin{figure}[t]
  \begin{minipage}[c]{0.56\textwidth}
    \centering
    \includegraphics[width=1\textwidth]{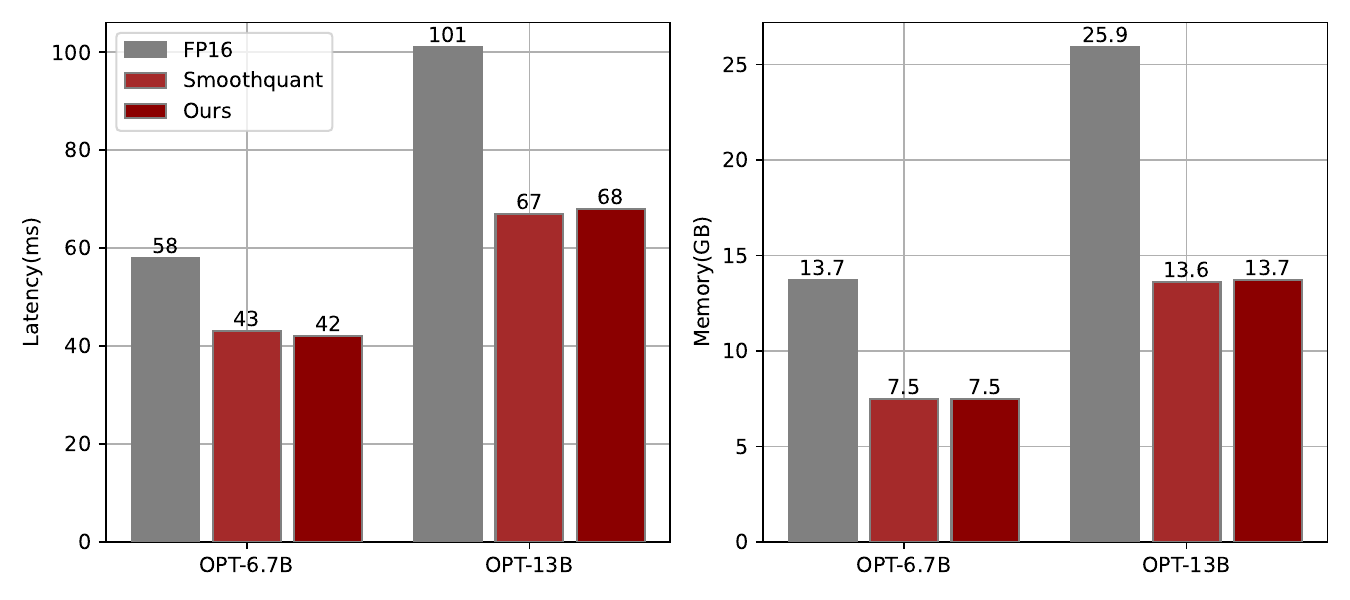}
    \caption{\label{fig3}Real latency and memory footprint of our proposed OutlierTune and Baseline. The batch size and sequence length are set to 1 and 512, respectively.}
  \end{minipage}
\hfill
  \begin{minipage}[c]{0.42\textwidth}
\captionsetup{type=table}
\caption{\label{tab5}
Real latency  of OutlierTune over different batch sizes on OPT Models. The sequence length is set to 128.}
\centering
\scalebox{0.81}{
\begin{tabular}{llllllllllllllllll}
\hline
  \multirow{2}{*}{{Models}}  &    \multirow{2}{*}{BS} & \multicolumn{3}{c}{{Latency (ms)}}  \\

\cmidrule(lr){3-5}
    &      & \makecell[c]{{FP16}} & \makecell[c]{{Ours}}  & \makecell[c]{{Speedup($\uparrow$)}}  \\

\hline
\multirow{2}{*}{{OPT-6.7B}}  &  \makecell[c]{{8}} & \makecell[c]{{12.1}} & \makecell[c]{8.2}  & \makecell[c]{1.48$\times$} \\

 &  \makecell[c]{{16}} & \makecell[c]{{11.2}} & \makecell[c]{8.1}  & \makecell[c]{1.38$\times$} \\

\hline 

\multirow{2}{*}{{OPT-13B}} &  \makecell[c]{{8}} & \makecell[c]{{21.9}} & \makecell[c]{15.6}  & \makecell[c]{1.41$\times$} \\

 &  \makecell[c]{{16}} & \makecell[c]{{19.1}} & \makecell[c]{14.3}  & \makecell[c]{1.34$\times$} \\

\hline 
\end{tabular}}
\end{minipage}
\end{figure}

In Fig. \ref{fig3}, we show the actual inference speed and memory usage of OutlierTune. 
Compared to FP16, OutlierTune exhibits remarkable improvements, with speedups of about 1.38x and 1.48x on OPT-6.7B and OPT-13B models, respectively. These performance gains are comparable to those observed with the Smoothquant method, with both techniques showing more efficient  as the scale of the model expands. Regarding memory utilization, both methods show approximately 2$\times$ savings.  In Table \ref{tab5}, we show that OutlierTune can significantly speed up the inference of LLMs across different batch sizes. Compared with FP16,  OutlierTune significantly reduces the per-token decoding latency, achieving a maximum improvement factor of 1.48$\times$.

\subsection{Ablation study}\label{sec4.4}
In this subsection, we investigate the effects of removing the symmetrization step prior to quantization. As shown in Table \ref{tab6}, the symmetrization operation only provides a marginal improvement under Int8 quantization. This observation suggests that Int8 quantization is sufficient to accurately represent the updated weights and activations under per-channel quantization. Moreover, advancing the quantization to Int6, the results with the symmetrization operation remain acceptable. In contrast, omitting symmetrization leads to a significant performance degradation.  This is due to the substantial variance between outlier and normal channels, which increases the disparity between different weights and heightens the challenge of weight quantization.

\begin{table}[tpb]
\caption{\label{tab6}
Effectiveness of the symmetrization operation.}
\centering
\scalebox{0.75}{
\begin{tabular}{llllllllllllllllll}
\hline
\multicolumn{1}{c}{{Model}}  & \multicolumn{4}{c}{{OPT-6.7B}}  & \multicolumn{4}{c}{{OPT-13B}}& \multicolumn{4}{c}{{OPT-30B}}\\
\cmidrule(lr){1-1}\cmidrule(lr){2-5}\cmidrule(lr){6-9}\cmidrule(lr){10-13}

 \multicolumn{1}{c}{{Task}} & \makecell[c]{{Wiki}} & \makecell[c]{{Ptb}} & \makecell[c]{{C4}} & \makecell[c]{{PIQA$\uparrow$}} & \makecell[c]{{Wiki}} & \makecell[c]{{Ptb}} & \makecell[c]{{C4}} & \makecell[c]{{PIQA$\uparrow$}} & \makecell[c]{{Wiki}} & \makecell[c]{{Ptb}} & \makecell[c]{{C4}} & \makecell[c]{{PIQA$\uparrow$}}\\

\hline
 \makecell[c]{Ours Int8}&   \makecell[c]{10.87}    &   \makecell[c]{13.10}   &  \makecell[c]{11.75}   &  \makecell[c]{76.39\%}  &   \makecell[c]{10.14}  &  \makecell[c]{12.34}  &  \makecell[c]{11.21}  &  \makecell[c]{75.63\%} &   \makecell[c]{9.57}   &  \makecell[c]{11.85}  &  \makecell[c]{10.70}  &  \makecell[c]{77.31\%}   \\

 \makecell[c]{Non-Sym Int8}&   \makecell[c]{10.93}    &   \makecell[c]{13.19}   &  \makecell[c]{11.78}   &  \makecell[c]{76.06\%}  &   \makecell[c]{10.15}  &  \makecell[c]{12.36}  &  \makecell[c]{11.22}  &  \makecell[c]{75.78\%} &   \makecell[c]{9.59}   &  \makecell[c]{11.86}  &  \makecell[c]{10.71}  &  \makecell[c]{77.91\%} \\

\makecell[c]{Ours Int6}&    \makecell[c]{11.49}    &   \makecell[c]{13.83}   &  \makecell[c]{12.22}   &  \makecell[c]{76.12\%}  &   \makecell[c]{11.78}  &  \makecell[c]{16.29}  &  \makecell[c]{12.43}  &  \makecell[c]{75.24\%} &   \makecell[c]{10.26}   &  \makecell[c]{13.30}  &  \makecell[c]{11.13}  &  \makecell[c]{76.77\%} \\

 \makecell[c]{Non-Sym Int6}&  \makecell[c]{6768}    &   \makecell[c]{6838}   &  \makecell[c]{10552}   &  \makecell[c]{51.47\%}  &   \makecell[c]{7077}  &  \makecell[c]{5531}  &  \makecell[c]{6109}  &  \makecell[c]{52.01\%} &   \makecell[c]{1348}   &  \makecell[c]{979}  &  \makecell[c]{934}  &  \makecell[c]{52.29\%} \\

\hline
\end{tabular}}
\end{table}

\section{Related Work}

\noindent\textbf{Quantization.} 
Quantization is a key compression method in reducing the computational complexity by converting floating-point values into fixed-point or low-precision floating-point representations \cite{nagel2020up}. This process effectively reduces the memory footprint of model parameters and speeds up the model inference.  Quantization techniques are typically divided into two main approaches: quantization-aware training (QAT) \cite{polino2018model,Jacob_2018_CVPR} and post-training quantization (PTQ) \cite{bengio2013estimating,choi2018pact}. QAT entails utilizing backpropagation to update quantized weights. However, this method incurs significant training costs, making it difficult to apply directly to LLMs \cite{10.5555/3524938.3525605}. In contrast, PTQ requires only a small number of samples and minimal resource consumption, facilitating  rapid completion of model quantization \cite{gholami2021survey}.

\noindent\textbf{Quantization of Transformer based LLMs.} 
Based on the difference of quantized objects, PTQ can be classified as: weight-only quantization \cite{wang2020towards,li2021brecq,sheng2023flexgen,dettmers2024qlora,lee2023owq,shen2020q} and weight-activation quantization \cite{luo2020positional,lin2020towards,choukroun2019low}. In weight-only quantization, model parameters are compressed, often to 4 bits or even lower, to reduce memory footprint and speed up inference \cite{frantar2022gptq,lin2023awq}. On the other hand,  weight-activation quantization  involves the  compression of both model weights and activations, typically to int8, offering potential acceleration through hardware-accelerated kernels \cite{wei-etal-2023-outlier,pmlr-v202-xiao23c}. This paper focuses on the  weight-activation quantization.

Recent studies have shown that the structured outliers in activations as the primary cause of activation quantization errors \cite{kovaleva2021bert}. Based on this finding, LLM.int8 \cite{dettmers2022llmint8} mitigates quantization errors by preserving outlier channels as FP16. Smoothquant \cite{pmlr-v202-xiao23c} optimizes int8 matrix multiplications by balancing the numerical ranges of activation and weight. Outlier suppression \cite{wei2022outlier} shows that scaling in LayerNorm expands the range of outlier channels, which can be mitigated by readjusting the scaling operation. QLLM \cite{liu2023qllm} employs outlier channel splitting to promote a more uniform distribution of activation numerical ranges. These methods are dedicated to the per-token or per-tensor activation quantization, which leads to higher quantization errors due to the presence of structured outliers.

Research by Xiao et al. \citep{pmlr-v202-xiao23c} indicates that the per-channel activation quantization can mitigate the dominant influence of outlier channels on the quantization scale. However, its dequantization process incurs a high additional computational burden, making it impractical for real applications. While methods like \cite{yuan2023rptq} and \cite{wu2023understanding} have been proposed to mitigate the computational overhead of per-channel quantization by directly reordering, they face challenges in achieving efficient hardware implementation and exhibit poor latency performance. 
We present OutlierTune, a novel weight-activation quantization framework that not only preserves the benefits of per-channel quantization but also addresses the computational inefficiencies inherent in current methods.

\section{Conclusion}
In this paper, we propose the OutlierTune framework for the efficient inference while maintaining the accuracy of per-channel activation quantization. By implementing the pre-execution of dequantization and symmetrization, OutlierTune efficiently performs the per-channel activation quantization and significantly mitigates the quantization errors during inference.  Experimental results show that OutlierTune is superior to existing methods on various tasks, improving the Int6 quantization of instruction-tuned LLM (OPT-IML) to the same level as FP16. In addition, it is 1.48x faster than FP16 with approximately 2x less memory usage.
This framework provides a viable and efficient solution for the deployment of LLMs in practical applications, promising broader and more effective utilization in future implementations.

\bibliographystyle{unsrt}
\bibliography{custom}

\newpage
\section*{NeurIPS Paper Checklist}

\begin{enumerate}

\item {\bf Claims}
    \item[] Question: Do the main claims made in the abstract and introduction accurately reflect the paper's contributions and scope?
    \item[] Answer: \answerYes{} 

\item {\bf Limitations}
    \item[] Question: Does the paper discuss the limitations of the work performed by the authors?
    \item[] Answer: \answerYes{} 
    \item[] Justification: In Appendix, we leave some topics as future work.

\item {\bf Theory Assumptions and Proofs}
    \item[] Question: For each theoretical result, does the paper provide the full set of assumptions and a complete (and correct) proof?
    \item[] Answer:  \answerYes{} 

    \item {\bf Experimental Result Reproducibility}
    \item[] Question: Does the paper fully disclose all the information needed to reproduce the main experimental results of the paper to the extent that it affects the main claims and/or conclusions of the paper (regardless of whether the code and data are provided or not)?
    \item[] Answer: \answerYes{} 

\item {\bf Open access to data and code}
    \item[] Question: Does the paper provide open access to the data and code, with sufficient instructions to faithfully reproduce the main experimental results, as described in supplemental material?
    \item[] Answer: \answerYes{} 
	 \item[] Justification: In Abstract, we give anonymous open source projects related to this work.

\item {\bf Experimental Setting/Details}
    \item[] Question: Does the paper specify all the training and test details (e.g., data splits, hyperparameters, how they were chosen, type of optimizer, etc.) necessary to understand the results?
    \item[] Answer: \answerYes{} 
\item[] Justification: Detailed setting can be found in the Section \ref{sec4}.

\item {\bf Experiment Statistical Significance}
    \item[] Question: Does the paper report error bars suitably and correctly defined or other appropriate information about the statistical significance of the experiments?
    \item[] Answer: \answerNA{} 

\item {\bf Experiments Compute Resources}
    \item[] Question: For each experiment, does the paper provide sufficient information on the computer resources (type of compute workers, memory, time of execution) needed to reproduce the experiments?
    \item[] Answer: \answerYes{} 
\item[] Justification: Detailed setting can be found in the Section \ref{sec4}.
    
\item {\bf Code Of Ethics}
    \item[] Question: Does the research conducted in the paper conform, in every respect, with the NeurIPS Code of Ethics \url{https://neurips.cc/public/EthicsGuidelines}?
    \item[] Answer: \answerYes{} 

\item {\bf Broader Impacts}
    \item[] Question: Does the paper discuss both potential positive societal impacts and negative societal impacts of the work performed?
    \item[] Answer: \answerNA{} 

\item {\bf Safeguards}
    \item[] Question: Does the paper describe safeguards that have been put in place for responsible release of data or models that have a high risk for misuse (e.g., pretrained language models, image generators, or scraped datasets)?
    \item[] Answer: \answerNA{} 

\item {\bf Licenses for existing assets}
    \item[] Question: Are the creators or original owners of assets (e.g., code, data, models), used in the paper, properly credited and are the license and terms of use explicitly mentioned and properly respected?
    \item[] Answer: \answerYes{} 

\item {\bf New Assets}
    \item[] Question: Are new assets introduced in the paper well documented and is the documentation provided alongside the assets?
    \item[] Answer: \answerNA{} 

\item {\bf Crowdsourcing and Research with Human Subjects}
    \item[] Question: For crowdsourcing experiments and research with human subjects, does the paper include the full text of instructions given to participants and screenshots, if applicable, as well as details about compensation (if any)? 
    \item[] Answer: \answerNA{} 

\item {\bf Institutional Review Board (IRB) Approvals or Equivalent for Research with Human Subjects}
    \item[] Question: Does the paper describe potential risks incurred by study participants, whether such risks were disclosed to the subjects, and whether Institutional Review Board (IRB) approvals (or an equivalent approval/review based on the requirements of your country or institution) were obtained?
    \item[] Answer: \answerNA{} 

\end{enumerate}

\newpage

\appendix
\section{Implementation details}\label{A}
To gather the essential statistics for pre-execution dequantization and symmetrization, we calibrate the symmetrization factors and determine static activation quantization parameters by analyzing 512 randomly selected sentences from the Pile dataset \cite{DBLP:journals/corr/abs-2101-00027} for zero-shot tasks. To achieve a balance between computational efficiency and model accuracy, we employ min-max calibration for weight quantizations in Int8* and all activation quantizations, along with the addition of zero points specifically for weight quantizations in Int8.  The simplicity of our approach obviates the need for intricate search algorithms, permitting the uniform application of our the pre-execution of dequantization and symmetrization framework across different models. This uniform application of our pre-execution dequantization and symmetrization framework allows for a comprehensive assessment of the generality and zero-shot efficacy of our proposed OutlierTune framework. All empirical evaluations are conducted using the PyTorch framework on 2 NVIDIA RTX A100 GPUs.

For the OPT and Bloom architectures, we note the implementation of residual connections prior to LayerNorm, an arrangement which allows us to substitute the post-LayerNorm linear layers with the mapping defined in \eqref{g} to establish symmetric equivalence. For Llama families \cite{touvron2023llama}, such models do not contain biases and cannot be symmetrized. Fortunately, our analysis reveals that their  activation outliers have less impact on the activation quantizations than Blooms and OPT models, and the outlier sizes vary across different tokens. Consequently, we apply Token-Wise Clipping as described in \citep{wei2022outlier} during calibration, enabling  accurate and efficient per-channel activation quantization.

\section{Algorithm efficiency}\label{B}
This paper investigates the quantization of both weights and activations through the introduction of the OutlierTune framework. OutlierTune does not necessitate re-training and seamlessly integrates with existing pre-trained models. In contrast to methods that involve the complicated parameter search process,  our framework is efficient. We evaluated the efficiency of OutlierTune by measuring the time required for calibration and full model quantization. As shown in Table \ref{tab7}, OutlierTune effectively computes symmetrization factors and static quantization parameters for an OPT-66B model in 9.3 minutes.   In addition, the complete quantization of the model parameters is accomplished in just 43.5 minutes. 
This rapid processing positions OutlierTune as a compelling solution for deploying LLMs in resource-constrained environments.

\begin{table}[h]
\caption{\label{tab7}
 OutlierTune runtime for full quantization of the 4 OPT models.}
\centering
\scalebox{0.9}{
\begin{tabular}{llllllllllllllllll}
\hline
 \multicolumn{1}{c}{{Models}}  & \makecell[c]{{OPT-6.7B}} & \makecell[c]{{OPT-13B}} & \makecell[c]{{OPT-30B}} & \makecell[c]{{OPT-66B}}  \\
\hline
 \makecell[c]{Calibration} & \makecell[c]{{1.9m}} & \makecell[c]{{2.8m}} &\makecell[c]{{4.1m}} &  \makecell[c]{{9.3m}} \\

 \makecell[c]{Runtime} & \makecell[c]{{4.3m}} & \makecell[c]{{8.6m}} &\makecell[c]{{19.6m}} &  \makecell[c]{{43.5m}} \\
\hline
\end{tabular}}
\end{table}

\section{Supplementary experiments}\label{D}

\paragraph{Results of Llama models.} In the Llama models, the absence of bias terms in each module precludes the use of symmetry mechanisms applied in other models. Fortunately, we found that the activation outliers in Llama exhibit a relatively small impact on activation quantization compared to other architectures.  We calibrate the static activation scaling factor through token-wise cropping, thereby mitigating the impact of the scaling factor on weights. As shown in Table \ref{tab8},  we show that OutlierTune quantizes the Llama2-7B to Int6 without incurring significant performance degradation.

\begin{table}[h] \vspace{0.4cm}
\caption{\label{tab8}
The  perplexity of our OutlierTune framework on Llama Models. }
\centering
\scalebox{0.9}{
\begin{tabular}{llllllllllllllllll} 
\hline
 \multicolumn{1}{c}{{Model}}  & \multicolumn{3}{c}{{Llama-2-7B}}  & \multicolumn{3}{c}{{Llama-3-8B}}\\
\cmidrule(lr){1-1}\cmidrule(lr){2-4}\cmidrule(lr){5-7}
 \multicolumn{1}{c}{{Tesk}}  & \makecell[c]{{FP16}} & \makecell[c]{{Int8}} & \makecell[c]{{Int6}}  & \makecell[c]{{FP16}} & \makecell[c]{{Int8}}   & \makecell[c]{{Int6}} \\
\hline
 \makecell[c]{WikiText2} & \makecell[c]{{5.47}} & \makecell[c]{{5.51}} & \makecell[c]{{6.15}}&  \makecell[c]{{6.14}} &  \makecell[c]{{6.26}} &  \makecell[c]{{7.70}}\\
 \makecell[c]{Ptb} & \makecell[c]{{20.83}} & \makecell[c]{{18.33}} &\makecell[c]{{22.23}} &  \makecell[c]{{10.59}} &  \makecell[c]{{10.75}} &  \makecell[c]{{12.57}}\\
 \makecell[c]{C4} & \makecell[c]{{11.07}} & \makecell[c]{{7.04}} & \makecell[c]{{7.79}} &  \makecell[c]{{8.88}} &  \makecell[c]{{9.04}} &  \makecell[c]{{6.15}}\\
\hline
\end{tabular}}\vspace{0.4cm}
\end{table}

\paragraph{Results of different LLMs.}
In Table \ref{tab9}, we provide a detailed evaluation of the perplexity performance of the Bloom-7B1 and OPT-IML-30B models. Our OutlierTune framework shows comparable or better perplexity results relative to the baselines across different quantization settings. In particular, in the case of 6-bit quantization, OutlierTune achieves significant perplexity improvements compared to the baselines. For example, OutlierTune improves perplexity scores by 27 and 12 on the Ptb task, respectively. Furthermore, the quantization of Bloom-7B1 is more challenging for the OutlierTune method,  as evidenced by a more pronounced performance degradation between the quantized Bloom model and the full-precision model compared to OPT-IML-30B.
\begin{table}[h]\vspace{0.4cm}
\caption{\label{tab9}
The  perplexity of our OutlierTune framework on Bloom-7B1 and OPT-IML-30B.}
\centering
\scalebox{0.9}{
\begin{tabular}{llllllllllllllllll}
\hline
 \multicolumn{1}{c}{{Bloom-7B1}}  & \makecell[c]{{FP16}} & \makecell[c]{{Smooth-Int8}} & \makecell[c]{{Ours-Int8}}  & \makecell[c]{{Smooth-Int6}} & \makecell[c]{{Ours-Int6}}  \\
\hline
 \makecell[c]{WikiText2} & \makecell[c]{{11.37}} & \makecell[c]{{11.51}} &\makecell[c]{\textbf{11.40}} &  \makecell[c]{{18.83}} &  \makecell[c]{\textbf{12.16}} \\
 \makecell[c]{Ptb} & \makecell[c]{{19.40}} & \makecell[c]{{19.79}} &\makecell[c]{\textbf{19.45}} &  \makecell[c]{{33.87}} &  \makecell[c]{\textbf{21.15}} \\
 \makecell[c]{C4} & \makecell[c]{{14.12}} & \makecell[c]{{14.29}} &\makecell[c]{\textbf{14.45}} &  \makecell[c]{{21.47}} &  \makecell[c]{\textbf{14.98}} \\
\hline
 \multicolumn{1}{c}{{OPT-IML-30B}}  & \makecell[c]{{FP16}} & \makecell[c]{{Smooth-Int8}} & \makecell[c]{{Ours-Int8}}  & \makecell[c]{{Smooth-Int6}} & \makecell[c]{{Ours-Int6}}  \\
\hline
 \makecell[c]{WikiText2} & \makecell[c]{{10.56}} & \makecell[c]{{10.56}} &\makecell[c]{\textbf{10.56}} &  \makecell[c]{{16.81}} &  \makecell[c]{\textbf{11.18}} \\
 \makecell[c]{Ptb} & \makecell[c]{{12.71}} & \makecell[c]{{12.74}} &\makecell[c]{\textbf{12.73}} &  \makecell[c]{{40.94}} &  \makecell[c]{\textbf{13.39}} \\
 \makecell[c]{C4} & \makecell[c]{{11.48}} & \makecell[c]{{11.51}} &\makecell[c]{\textbf{11.49}} &  \makecell[c]{{26.43}} &  \makecell[c]{\textbf{11.78}} \\
\hline
\end{tabular}}
\end{table}\vspace{0.4cm}

\section{Combination with the GPTQ}\label{C}
We conducted a detailed analysis of the quantification difficulty distribution of OutlierTune, and the specific results are shown in Table \ref{tab10}. Our analysis revealed a significant  change in the difficulty of quantizing both weights and activations after implementing the pre-execution of dequantization and symmetrization. The modifications in the weights emerged as the primary contributors to quantization errors. Notably, our results show that, by keeping weight parameters constant, it is possible to quantize activations to Int4 with minimal quantization losses. This highlights the potential for optimizing quantization strategies to improve performance while minimizing performance degradation.

\begin{table}[h]\vspace{0.4cm}
\caption{\label{tab10}
Quantitative difficulty analysis of weights and activations under the OutlierTune framework.}
\centering
\scalebox{0.9}{
\begin{tabular}{llllllllllllllllll}
\hline
\makecell[c]{{Tesk}}   & \makecell[c]{{FP16}} & \makecell[c]{{W8A6}} & \makecell[c]{{W6A6}} & \makecell[c]{{W8A4}} & \makecell[c]{{W4A4}}\\
\hline
 \makecell[c]{WikiText2} & \makecell[c]{{10.86}} & \makecell[c]{{10.95}} & \makecell[c]{{11.49}} &  \makecell[c]{{15.95}} &  \makecell[c]{{810.39}}\\
 \makecell[c]{Ptb} & \makecell[c]{{13.09}} & \makecell[c]{{13.20}} & \makecell[c]{{13.83}} &  \makecell[c]{{14.42}} &  \makecell[c]{{1102.17}}\\
 \makecell[c]{C4} & \makecell[c]{{11.74}} & \makecell[c]{{11.82}} & \makecell[c]{{12.22}} &  \makecell[c]{{15.99}} &  \makecell[c]{{1260.09}}\\
\hline
\end{tabular}}\vspace{0.4cm}
\end{table}

Based on the above analysis, we combine the OutlierTune and GPTQ methods. GPTQ quantizes the model weights based on a second-order approximation, and the weights can be quantized to Int3 without excessive loss in accuracy. Table \ref{tab11} illustrates the outcomes of combining the OutlierTune and GPTQ techniques. The results show that our approach complements GPTQ, thereby improving the performance of W6A6 quantization. Nonetheless, a significant performance degradation is observed with W4A4 quantization. We leave this problem as the focus of our future work.
\begin{table}[h]\vspace{-0.2cm}
\caption{\label{tab11}
When combined with GPTQ, the performance gap under W6A6 is further reduced.}
\centering
\scalebox{0.9}{
\begin{tabular}{llllllllllllllllll}
\hline
\makecell[c]{{Tesk}}   & \makecell[c]{{FP16}} & \makecell[c]{{Our-Int6}} & \makecell[c]{{(Our+GPTQ)-Int6}} & \makecell[c]{{Our-Int4}} & \makecell[c]{{(Our+GPTQ)-Int4}}\\
\hline
 \makecell[c]{WikiText2} & \makecell[c]{{10.86}} & \makecell[c]{{11.49}} & \makecell[c]{{11.08}} &  \makecell[c]{{810.39}} &  \makecell[c]{{691.45}}\\
 \makecell[c]{Ptb} & \makecell[c]{{13.09}} & \makecell[c]{{13.83}} & \makecell[c]{{13.41}} &  \makecell[c]{{1102.17}} &  \makecell[c]{{881.09}}\\
 \makecell[c]{C4} & \makecell[c]{{11.74}} & \makecell[c]{{12.22}} & \makecell[c]{{12.00}} &  \makecell[c]{{1260.09}} &  \makecell[c]{{1032.06}}\\
\hline
\end{tabular}}\vspace{-0.2cm}
\end{table}

\end{document}